\documentclass[10pt,twocolumn,letterpaper]{article}

\usepackage{wacv}
\usepackage{times}
\usepackage{epsfig}
\usepackage{graphicx}
\usepackage{amsmath}
\usepackage{amssymb}
\usepackage{booktabs}
\usepackage{array}
\usepackage[accsupp]{axessibility}  

\def\wacvPaperID{199} 

\wacvalgorithmstrack   

\wacvfinalcopy 

\ifwacvfinal
\usepackage[breaklinks=true,bookmarks=false]{hyperref}
\else
\usepackage[pagebackref=true,breaklinks=true,colorlinks,bookmarks=false]{hyperref}
\fi

\begin{document}

\title{Cut-Paste Consistency Learning for Semi-Supervised Lesion Segmentation}

\author{Boon Peng Yap, Beng Koon Ng\\
School of Electrical and Electronic Engineering\\
Nanyang Technological University, Singapore\\
{\tt\small \{boonpeng001, ebkng\}@ntu.edu.sg}
}

\maketitle
\thispagestyle{empty}

\begin{abstract}
    Semi-supervised learning has the potential to improve the data-efficiency of training data-hungry deep neural networks, which is especially important for medical image analysis tasks where labeled data is scarce. In this work, we present a simple semi-supervised learning method for lesion segmentation tasks based on the ideas of cut-paste augmentation and consistency regularization. By exploiting the mask information available in the labeled data, we synthesize partially labeled samples from the unlabeled images so that the usual supervised learning objective (e.g., binary cross entropy) can be applied. Additionally, we introduce a background consistency term to regularize the training on the unlabeled background regions of the synthetic images. We empirically verify the effectiveness of the proposed method on two public lesion segmentation datasets, including an eye fundus photograph dataset and a brain CT scan dataset. The experiment results indicate that our method achieves consistent and superior performance over other self-training and consistency-based methods without introducing sophisticated network components.
\end{abstract}

\section{Introduction}
\begin{figure}
\begin{center}
\includegraphics[height=236pt]{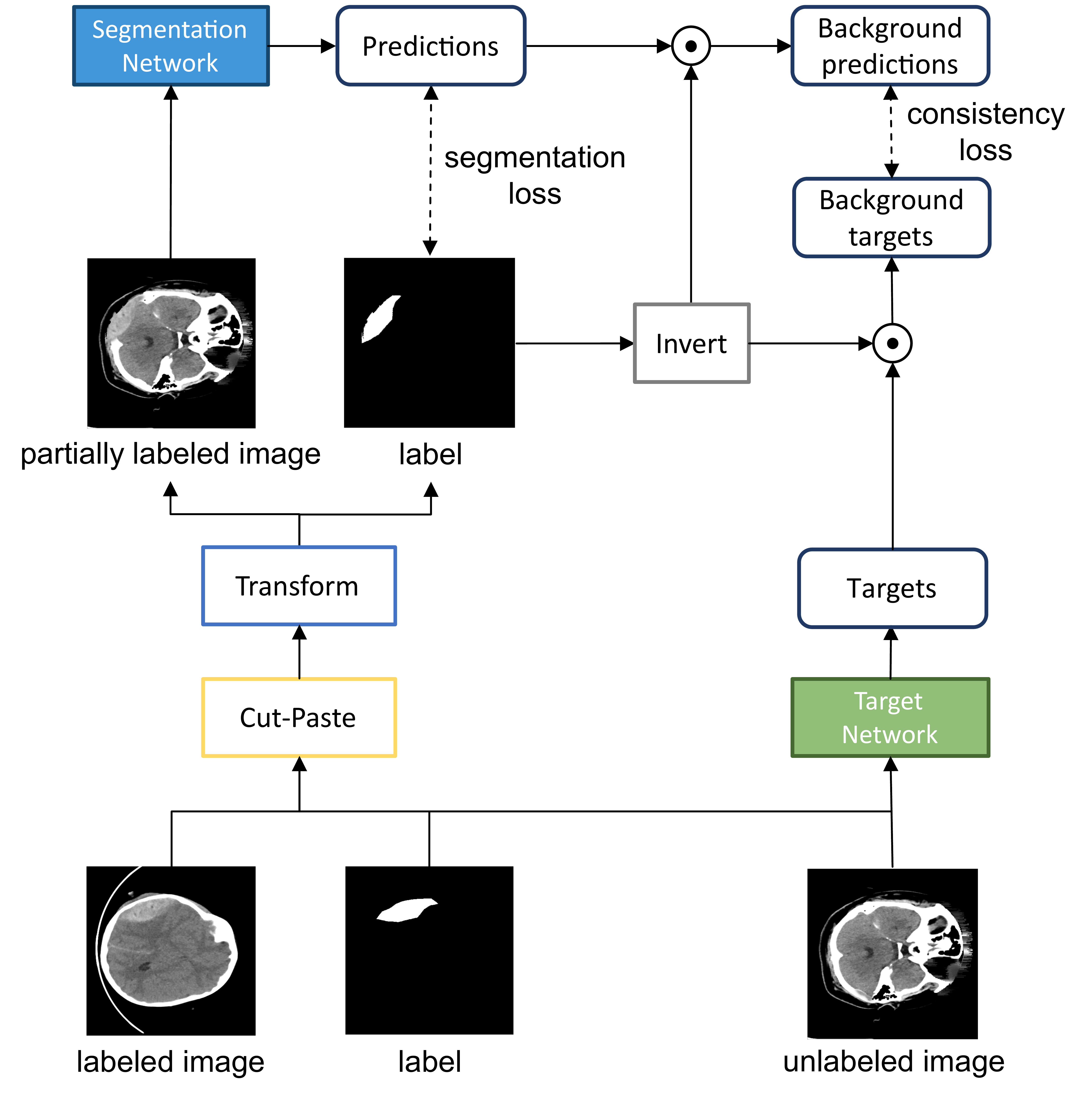}
\end{center}
\caption{Illustration of the proposed method. Partially labeled samples are generated by blending the foreground objects from labeled samples with unlabeled images. For binary segmentation tasks, predictions and targets are probability maps of the positive class. Target network can be an exact copy of the segmentation network or an exponential moving average version of the network. $\odot$ denotes element-wise multiplication.}
\label{method}
\end{figure}
Deep neural networks are known to be data-hungry – a large set of labeled training examples are required to generalize well to unseen testing data. In the medical imaging domain, labels are prohibitively expensive to acquire as it requires domain-specific knowledge from professionally trained specialists. To relieve the burden of label collection, data-efficient learning methods such as those based on semi-supervised learning are being actively pursued. Semi-supervised learning enables joint learning on both labeled and unlabeled examples, and the resulting model performs better than just training on the labeled examples alone. Semi-supervised learning methods can be broadly categorized into three categories: 1) pseudo-labeling \cite{lee2013pseudo, zoph2020rethinking, Shi2021InconsistencyawareUE, Kamraoui2021POPCORNPP}, 2) consistency learning \cite{tarvainen2017mean, NEURIPS2020_44feb009, french2020semi, Olsson2021ClassMixSD, peng2020deep, Ouali2020SemiSupervisedSS, Kamraoui2021POPCORNPP}, and 3) auxiliary task learning \cite{kervadec2019curriculum, zhai2019s4l}. Pseudo-labeling utilizes pretrained models to generate pseudo-labels for the unlabeled data, consistency learning encourages different perturbed views of the same input to have similar feature representations or outputs, while auxiliary task learning makes use of proxy tasks derived from the unlabeled data. Despite obtaining impressive performance gains, many methods rely on adding sophisticated components to the neural networks, which introduce a sizeable computation overhead to the training process. In this work, a cut-paste consistency-based semi-supervised learning method is proposed for lesion segmentation tasks. This method does not introduce any computationally intensive component and is specifically tailored for segmentation of small irregular objects such as lesions. Compared to organ segmentation, lesions have much larger variations in size, position and texture, which makes them harder to delineate. Based on the observation that lesions are small and sparsely distributed in images with much higher resolutions, the main idea is to treat the unlabeled images as a set of lesion-free backgrounds and use the mask information in labeled data as a source of foreground objects to synthesize new training examples, a process related to the cut-paste augmentation technique.

Cut-paste, also known as Copy-paste, is an augmentation technique unique to image detection and segmentation tasks. This augmentation technique extracts foreground objects using existing bounding boxes or segmentation masks and pastes them on random images to create a more diverse training set. Cut-paste has also been applied successfully to the semi-supervised setting in a self-training setup \cite{Ghiasi2021SimpleCI}, in which labeled masks are randomly pasted onto the pseudo-labeled masks. Unlike the self-training setup which requires at least two rounds of training (first round – train on the labeled data; second round – train on mixture of labeled and pseudo-labeled data), our proposed method only require one round of training and does not rely on fixed pseudo-labels.

The main contribution in this work is three-fold: 1) we demonstrated how to construct partially labeled samples for tiny objects segmentation (e.g., lesions) using unlabeled images and the mask information of labeled data; 2) we formulated a simple but effective training objective for joint learning on the pool of labeled and partially labeled samples; 3) we conducted experiments to verify the effectiveness of the proposed method on two public benchmarks for lesion segmentation. \footnote{Codes are made available at https://github.com/BPYap/Cut-Paste-Consistency.}

\section{Literature Review}
This section reviews related work on consistency-based semi-supervised learning and cut-paste augmentations, the two important ingredients in the proposed method.

\subsection{Consistency-based Semi-Supervised Learning}
Many popular semi-supervised learning methods are based on consistency regularization which aims to learn noise-invariant representations from unlabeled samples by minimizing the discrepancy in the output distributions of different perturbed views of the same input sample. To generate diverse perturbations, Virtual Adversarial Training (VAT) \cite{miyato2018virtual} perturbs each sample towards an adversarial direction that results in the largest change in output distribution while Unsupervised Data Augmentation (UDA) \cite{NEURIPS2020_44feb009} perturbs the samples with stronger and more realistic data augmentation strategies. For semi-supervised semantic segmentation, different mixing-based perturbation strategies have been explored, including an approach \cite{french2020semi} of generating pseudo-targets by mixing the predictions of two unlabeled samples using CutMix \cite{Yun2019CutMixRS}. Another approach called ClassMix \cite{Olsson2021ClassMixSD} generates pseudo-targets with more refined object boundaries by converting the predicted segmentation maps into binary masks prior to mixing. More recent methods \cite{Alonso2021SemiSupervisedSS, Zhou2021C3SemiSegCS, Zhong2021PixelCS, wang2022semi} complement consistency regularization with pixel-wise contrastive learning. Unlike these methods which only operate on unlabeled samples, the proposed method in this work synthesizes partially labeled samples from the labeled samples to improve the diversity of training samples.

\subsection{Cut-paste Augmentation}
First introduced for instance detection \cite{Dwibedi2017CutPA}, cut-paste has been studied extensively in tasks like instance segmentation \cite{Ghiasi2021SimpleCI} and semantic segmentation \cite{CDA_2021_ICCV, yang2021tumorcp, Tranheden2021DACSDA, Cardace2022PluggingSM}. The idea is simple – first, foreground objects are extracted from one image using previously annotated bounding boxes or segmentation masks. Then, the extracted objects are blended with other image to create a new training sample. To prevent neural networks from picking up shortcuts/noises caused by the pasting operation, several works have been introduced to generate more realistic images by modelling the context of the backgrounds. For example, Cut\&Paste \cite{remez2018learning} pastes objects on the same horizontal scanline to preserve the correct scale after pasting, Context-DA \cite{dvornik2018modeling} trains a context model to select the most likely background position for pasting, while InstaBoost \cite{Fang2019InstaBoostBI} selects pasting locations according to the appearance consistency heatmaps computed from handcrafted descriptors. Since medical images within the same modality have similar appearance, previously proposed criteria for modelling the contexts of natural images might not be applicable to medical images without substantial modifications. In this work, a simpler alternative based on measuring the similarity of pixel values is explored.  

\section{Method}

\begin{figure*}
\begin{center}
\includegraphics[width=430pt]{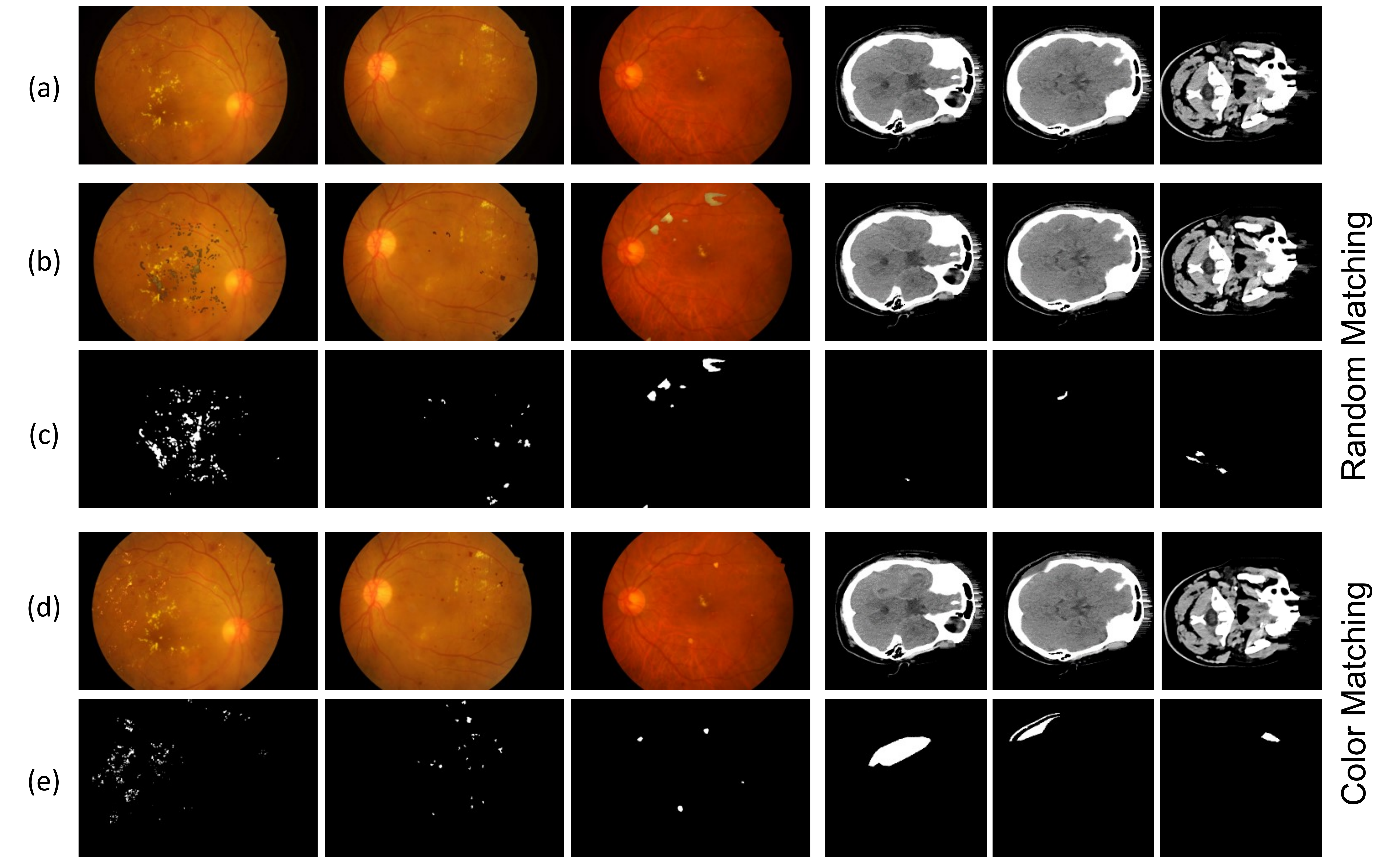}
\end{center}
\caption{(a) Three fundus images and three CT scans randomly sampled from unlabeled datasets. (b) Synthetic images and (c) labels generated by cutting and pasting random lesions from the labeled dataset. (d) Synthetic images and (e) labels generated by cutting and pasting lesions from labeled images with similar colours (for fundus images) or grayscale intensities (for CT scans).}
\label{samples}
\end{figure*}

Similar to CutMix and ClassMix, this work focuses on mixing-based perturbation strategy for semi-supervised semantic segmentation. The core idea of the proposed cut-paste consistency learning is as follow: utilizing the mask information available in the labeled data, a set of synthetic samples are generated by cutting and pasting masked regions from the labeled data onto the unlabeled images. This converts the unlabeled images into partially labeled synthetic samples and allows the cross entropy function to be used as training objective. This is useful for learning lesion boundaries under different background conditions. Since the synthetic samples are partially labeled (i.e., regions outside the pasted objects are still considered as unlabeled), the cross entropy objective will incorrectly penalize the unlabeled regions that are actually positive. Thus, to account for the uncertain labels in the backgrounds of the synthetic samples, a background consistency term is added to regularize the loss function. This consistency term encourages similar outputs in the backgrounds for both the original images (before cut-paste) and synthetic images as their predictions vary during the course of training. An illustration of the proposed method is shown in Fig. \ref{method}. The rest of this section describes the details of the image synthesis process, followed by a description of background consistency regularization and a formulation of the overall loss function. 

\subsection{Image Synthesis}
Most existing semi-supervised learning methods indirectly propagate label information to the unlabeled data through a joint optimization of separate loss terms. Our proposed cut-paste consistency method explicitly creates a direct link between the labeled and unlabeled data through an image synthesis process that pastes masked regions from the labeled images onto the unlabeled images. This approach allows known foreground objects (e.g., lesions) to be incorporated into a wide variety of background scenes in the unlabeled images to create a more diverse training data for better model generalization. To synthesize a new sample, each unlabeled image is first matched with a labeled sample (an image-mask pair). Foreground objects are extracted from the labeled image with the object mask and transformed before blending with the unlabeled image. The blended image, along with its mask, are added to the labeled dataset and is used as the training set for semi-supervised learning.

\paragraph{Color Matching} When matching unlabeled images with labeled samples, simply choosing a random sample usually results in a synthesized image that are visually inconsistent, making it too trivial for the segmentation network to learn meaningful features. To achieve a more realistic synthetic image, the samples are matched according to their pixel value similarities. For color images, unlabeled images are matched with labeled images with the lowest Delta E distances (a measure of color difference in the CIELAB color space); for grayscale images, the samples are matched based on the lowest L2 distance in gray-level pixel intensity. Some synthetic samples generated using the random and color matching scheme are shown in Fig. \ref{samples} for comparison. From the figure, the fundus images synthesized via the color matching scheme appear to be more realistic compared to those generated with random matching. For the CT scans, the color matching scheme is more likely to select the appropriate lesions by matching the unlabeled images with similarly-sized scans. During training, the synthetic samples are generated on the fly and each unlabeled image is paired with a randomly selected sample from the five most similar labeled samples.

\begin{figure}
\begin{center}
\includegraphics[height=200pt]{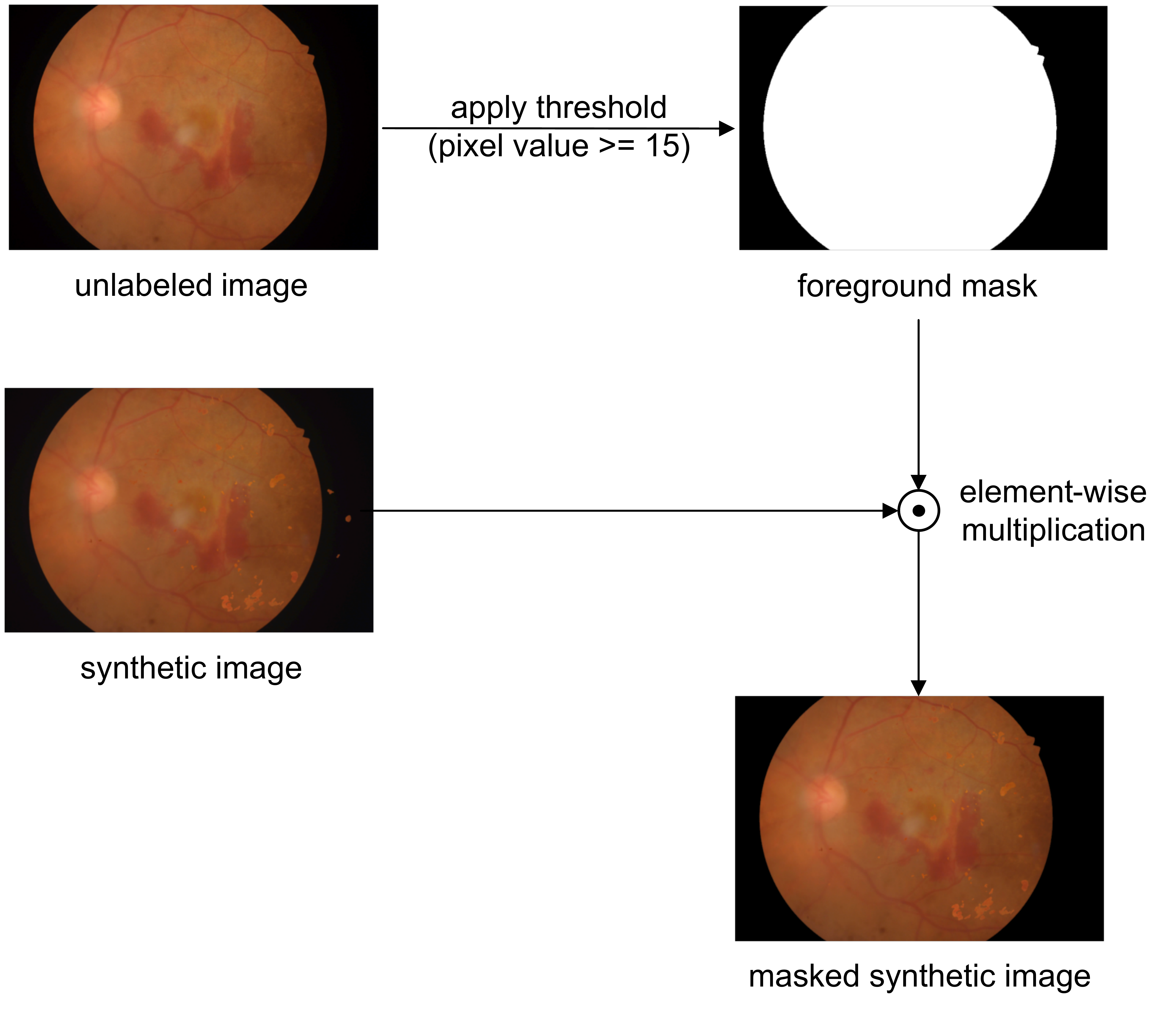}
\end{center}
\caption{Post-processing step to remove objects pasted on out-of-bound regions.}
\label{clip}
\end{figure}

\paragraph{Image Blending} To increase diversity in the synthetic samples, small amount of geometric and color jitterings are added to the foreground objects before pasting onto the unlabeled images. The geometric jittering consists of random rotation, random translation, and random resizing while the color jittering includes random jittering in the brightness, contrast, and saturation level. The transformed foreground objects are then blended with the unlabeled images via Gaussian blurring. We found that it is also beneficial to add a small amount of Gaussian noise to the unlabeled images prior to image blending. This also acts as a source of noise for background consistency regularization. A post-processing step is applied to mask out any possible out-of-bound objects from the blended images. As illustrated in Fig. \ref{clip}, this is achieved using image-specific binary masks obtained by thresholding each un-blended image with the pixel value of the out-of-bound regions.

\subsection{Background Consistency Regularization}
In the synthesized images, regions outside of the pasted objects are still considered as unlabeled. To prevent incorrect penalization of the potential false-negative regions by the cross entropy objective, a consistency regularization term is applied to the background predictions. Specifically, the background prediction of each synthetic image is obtained as the result of an element-wise multiplication between the network prediction and the inverted synthetic mask. The original unlabeled image is passed through a target network to obtain a background target, which is then multiplied by the same inverted synthetic mask used in the computation of the background prediction. When computing the background target, a stop gradient operation is applied to prevent the weights of the target network from being updated. This target network can either be an exact copy of the segmentation network being optimized, or an exponential moving average version of the network \cite{tarvainen2017mean}. The latter tends to produce a more stable target for consistency regularization. The background prediction is then encouraged to be consistent with the background target by minimizing the mean squared error of the output distributions:
\begin{equation}
    \mathcal{L}_{bg}(x, \tilde{x}, \tilde{y}) = \left\lVert f(\tilde{x}) \odot (1 - \tilde{y}) - f'(x) \odot (1 - \tilde{y})\right\rVert^{2}_{2}
\end{equation}
where \(f\) and \(f'\) are the segmentation network and target network respectively, \(x\), \(\tilde{x}\) and \(\tilde{y}\) are the original (unlabeled) images, synthetic images and synthetic segmentation masks respectively.

\subsection{Overall Loss Function}
Combined with the labeled term, the overall loss function is given by:
\begin{align}
    L      ={}& L_\ell + \lambda_u L_u \\
    L_\ell ={}& \sum_{(x, y) \in \mathcal{D}_\ell}{\mathcal{L}_{bce}(f(x), y)} \\
    {L}_u  ={}& \sum_{(x, \tilde{x}, \tilde{y}) \in \mathcal{D}_s}[\mathcal{L}_{bce}(f(\tilde{x}), \tilde{y}) + \mathcal{L}_{bg}(x, \tilde{x}, \tilde{y})]
\end{align}
where \(\mathcal{D}_\ell\) and \(\mathcal{D}_s\) are the labeled dataset and synthetic dataset respectively, and \(\lambda_u\) is a task-specific hyperparameter controlling the contribution of the unlabeled loss term. Notably, the unlabeled term (\(L_u\)) takes additional synthetic samples as inputs, in contrast to prior semi-supervised learning objectives which only takes unlabled samples as inputs. \(\mathcal{L}_{bce}\) represents the (weighted) binary cross entropy loss. As most lesion segmentation datasets are heavily imbalanced, more weights are given to the positive instances in \(\mathcal{L}_{bce}\). Specifically, the positive weight is calculated as:
\begin{equation}
    w_{pos} = \ln{\frac{P_{total}}{P_{pos}}}
\end{equation}
where \(P_{total}\) refers to the total number of pixels in all images, and \(P_{pos}\) is the number of pixels labeled as positive.

\section{Experiments and Results}
This section presents the details and results of semi-supervised learning experiments on two public benchmarks for lesion segmentation. Additionally, ablation studies are conducted to study the impact of each component in the proposed method.

\subsection{Datasets}
The effectiveness of the proposed method is evaluated on two publicly available datasets for lesion segmentation: the Indian Diabetic Retinopathy Image Dataset (IDRiD)\footnote{https://idrid.grand-challenge.org/Home/} \cite{porwal2020idrid} and a computed tomography (CT) dataset for intracranial hemorrhage segmentation (CT-ICH)\footnote{https://physionet.org/content/ct-ich/1.3.1/} \cite{hssayeni2020intracranial}.

IDRiD is a challenge dataset consisting of up to 81 color fundus images with pixel-level annotation for four types of retinal lesions: microaneurysms (81 images), hemorrhage (80 images), hard exudates (81 images), and soft exudates (40 images). It also contains 435 other images with image-level annotation for diabetic retinopathy and diabetic macular edema. For semi-supervised lesion segmentation, these 435 images are treated as unlabeled data.

The CT-ICH dataset contains 82 CT scans, of which 36 of them are collected from patients diagnosed with intracranial hemorrhage. Each CT scan includes about 30 slices with 5 mm slice-thickness, and regions with intracranial hemorrhage were manually delineated by two radiologists. In total, 2814 slices were extracted from the CT scan data. Unlike IDRiD where each fundus image from the labeled data is guaranteed to contain at least one type of lesion, most of the slices in the CT-ICH dataset does not contain intracranial hemorrhage. This makes it a heavily imbalanced dataset and extra challenging for semi-supervised learning. To simulate the scenario of semi-supervised learning, a portion of the full dataset are chosen as labeled data via patient-wise stratified sampling. The rest of the dataset is treated as unlabeled data.

\begin{table*}[!ht]
\begin{center}
\begin{tabular}{l>{\centering}p{35pt}>{\centering}p{35pt}>{\centering}p{35pt}>{\centering}p{35pt}>{\centering\arraybackslash}p{60pt}}
\toprule
 \multicolumn{1}{l}{Method} & MA & HE & EX & SE & Average  \\
\midrule
 Supervised (labeled data only)                      &  48.64               &   62.40               &   82.95               &   73.82               &   66.95 \\
 \midrule
 Self-training \cite{zoph2020rethinking}             &  49.57	            &   65.66           	&   86.08           	&   73.96	            &   68.82 \\  
 Self-training + Cut-Paste \cite{Ghiasi2021SimpleCI} &  49.63	            &   65.84	            &   86.92           	&   73.86	            &   69.06 \\
 Mean Teacher* \cite{tarvainen2017mean}              &  50.37	            &   64.71	            &   83.23           	&   75.86	            &   68.54 \\
 CutMix consistency* \cite{french2020semi}           &  \textbf{51.11}	    &   63.10	            &   85.17	            &   72.13	            &   67.88 \\
 ClassMix consistency* \cite{Olsson2021ClassMixSD}   &  42.29	            &   63.63	            &   87.17	            &   76.29	            &   67.34 \\
 \midrule
 Cut-Paste consistency (this work)                        &  50.20	            &   \textbf{66.67}	    &   \underline{87.24}	&   \underline{76.91}	&   \underline{70.26} \\
 Cut-Paste consistency* (this work)                       &  \underline{51.03}	&   \underline{65.92}	&   \textbf{88.47}	    &   \textbf{78.20}	    &   \textbf{70.91} \\
\bottomrule
\end{tabular}
\end{center}
\caption{Comparison of AUC-PR scores (\%) on four lesion segmentation tasks in IDRiD. Best result is shown in \textbf{bold}, and the second best result is \underline{underlined}. *Method that uses Mean Teacher model \cite{tarvainen2017mean} in consistency learning. (Abbreviations: MA - microaneurysms; HE - hemorrhage; EX - hard exudates; SE - soft exudates)}
\label{tab_idrid}
\end{table*}

\subsection{Experiment Setups}
\paragraph{Baselines} The proposed cut-paste consistency learning method is benchmarked against six baseline methods, including a fully supervised (i.e., training on the available labeled data only) baseline and five prior semi-supervised learning methods: (i) self-training \cite{zoph2020rethinking}, (ii) self-training + cut-paste \cite{Ghiasi2021SimpleCI}, (iii) Mean Teacher \cite{tarvainen2017mean}, (iv) CutMix consistency \cite{french2020semi}, and (v) ClassMix consistency \cite{Olsson2021ClassMixSD}. To standardize the benchmarking process and ensure fair comparisons, all semi-supervised methods are re-implemented in a shared codebase using the Pytorch \cite{NEURIPS2019_9015} library.

\paragraph{Data Augmentations} Each image along with its segmentation mask is resized to 512 pixels along the shorter side. During training, standard augmentations such as random rotation, random flipping, and random grayscaling (only applicable for IDRiD) are applied to each input sample. During inference, the predicted segmentation masks are upscaled to their original sizes.

\begin{table*}[ht]
\begin{center}
\begin{tabular}{lcccccccc}
\toprule
  & \multicolumn{2}{c}{30\%} & \multicolumn{2}{c}{50\%} & \multicolumn{2}{c}{70\%} & \multicolumn{2}{c}{100\%}  \\
\cmidrule(l){2-3}
\cmidrule(l){4-5}
\cmidrule(l){6-7}
\cmidrule(l){8-9}
\multicolumn{1}{l}{Method} & F1 & Jacc. & F1 & Jacc. & F1 & Jacc. & F1 & Jacc. \\
\midrule
 Supervised (labeled data only)                      &  23.93             &	14.39             &	37.65             &	24.38             &	46.92             &	30.80 &	55.71     &	39.50 \\
 \midrule
 Self-training \cite{zoph2020rethinking}             &  27.49             & 16.52             &	40.88             &	26.60             &	48.18             &	32.41             &	- &	- \\
 Self-training + Cut-Paste \cite{Ghiasi2021SimpleCI} &  \underline{32.59} &	\underline{21.28} &	\textbf{43.00}    &	\underline{28.47} &	48.52             &	32.43             &	- &	- \\
 Mean Teacher* \cite{tarvainen2017mean}              &  24.70             &	14.78             &	39.61             &	25.75             &	48.26             &	32.21             &	- &	- \\
 CutMix consistency* \cite{french2020semi}           &  27.02             &	16.70             &	39.89             &	26.10             &	47.87             &	32.35             &	- &	- \\
 ClassMix consistency* \cite{Olsson2021ClassMixSD}   &  15.01             &	8.77              &	33.54             &	21.50             &	\underline{49.51} &	\underline{33.37} &	- &	- \\
 \midrule
 Cut-Paste consistency* (this work)                       &  \textbf{35.68}    &	\textbf{23.94}    &	\underline{42.55} &	\textbf{28.69}    &	\textbf{49.82}    &	\textbf{33.38}    &	- &	- \\
\bottomrule
\end{tabular}
\end{center}
\caption{Comparison of F1 scores (\%) and Jaccard indices (\%) on the CT-ICH dataset with different ratios of labeled examples. 100\% labeled setting represents the performance upper bound achievable using all available labels. Best result is shown in \textbf{bold}, and the second best result is \underline{underlined}. *Method that uses Mean Teacher model \cite{tarvainen2017mean} in consistency learning.}
\label{tab_ich}
\end{table*}

\begin{table*}[ht]
\begin{center}
\begin{tabular}{>{\centering}p{50pt}>{\centering}p{50pt}>{\centering}p{50pt}>{\centering}p{50pt}>{\centering}p{50pt}>{\centering}p{40pt}>{\centering}p{35pt}>{\centering\arraybackslash}p{35pt}}
\toprule
 \begin{tabular}[c]{@{}c@{}}Mask\\ Blurring\end{tabular} & \begin{tabular}[c]{@{}c@{}}Background\\ Blurring\end{tabular} & \begin{tabular}[c]{@{}c@{}}Color\\ Matching\end{tabular} & \begin{tabular}[c]{@{}c@{}}Image\\ Consistency\end{tabular} & \begin{tabular}[c]{@{}c@{}}Background\\ Consistency\end{tabular} & \(\lambda_{u}\) & EX & MA  \\
\midrule
            &            &            &            &                  & 0.01 & 81.27 & 49.85 \\
 \checkmark &            &            &            &                  & 0.01 & 82.35 & 49.78 \\
            & \checkmark &            &            &                  & 0.01 & 82.60 & 49.12 \\
 \checkmark & \checkmark &            &            &                  & 0.01 & 84.31 & 48.90 \\
 \checkmark & \checkmark & \checkmark &            &                  & 0.01 & 85.24 & 49.29 \\
 \checkmark & \checkmark & \checkmark & \checkmark &                  & 0.01 & 85.86 & 50.85 \\
 \checkmark & \checkmark & \checkmark &            & CE  & 0.01 & 88.24 & 50.31 \\
 \checkmark & \checkmark & \checkmark &            & MSE & 0.01 & \textbf{88.47} & \textbf{51.03} \\
\midrule
 \checkmark & \checkmark & \checkmark &            & MSE & 0.009 & 86.60 & 50.50 \\
 \checkmark & \checkmark & \checkmark &            & MSE & 0.03  & 88.37 & 49.83 \\
 \checkmark & \checkmark & \checkmark &            & MSE & 0.05  & 87.03 & 50.90 \\
\bottomrule
\end{tabular}
\end{center}
\caption{Results of ablation studies (AUC-PR scores) on hard exudates (EX) and microaneurysms (MA) segmentation tasks in IDRiD.}
\label{tab_ablation}
\end{table*}

\paragraph{Training Details} The popular U-Net \cite{ronneberger2015u} is chosen as the segmentation network. For each task, the weight decay value is fixed at \(10^{-5}\) and the learning rate is tuned on the fully supervised baseline only. The selected learning rates are kept fixed for all the semi-supervised learning experiments. Each segmentation network is optimized using the AdamW optimizer \cite{Loshchilov2019DecoupledWD} for up to 500 epochs and 50 epochs for the IDRiD and CT-ICH datasets, respectively. 10\% of the training samples are randomly selected as validation dataset and early stopping is applied based on the performance on the validation dataset. During training, the learning rate is linearly warmup in the first 10 epochs and gradually decayed after the 10-th epoch using a cosine scheduler. The batch size for the IDRiD and CT-ICH dataset is set to be 5 and 8, respectively. The task-specific weighting parameters (\(\lambda_u\)) for cut-paste consistency learning are empirically set to 0.01 for all tasks in IDRiD and 0.1 for intracranial hemorrhage segmentation in the CT-ICH dataset. For lesion segmentation with IDRiD, training is repeated with different random seeds on the training set and the average performance measure on the testing set is reported. For hemorrhage segmentation with the CT-ICH dataset, five-fold cross-validations are performed. All segmentation networks are optimized using a single NVIDIA V100 GPU with 16 GB memory.

\subsection{Results on IDRiD} Table \ref{tab_idrid} shows the evaluation results on the test set of each lesion segmentation task in IDRiD. For this dataset, the segmentation performance is measured using area under the precision-recall curve (AUC-PR), following the official performance measure from the challenge organizer. From the table, the proposed cut-paste consistency learning method consistently places among the top performing methods and achieves the best performance in hemorrhage (HE) segmentation. Through a straightforward extension with the Mean Teacher model \cite{tarvainen2017mean} (i.e., replacing the target network with the exponential moving average version of the segmentation network), our method achieved the best average result across the four lesion segmentation tasks, with 3.96\% improvement over the fully supervised baseline and a margin of 1.85\% compared to the best baseline method (self-training + cut-paste). Self-training, with or without cut-paste, is a competitive and simple-to-implement semi-supervised method. However, it requires longer training time compared to other methods, since multiple rounds of trainings is typically needed to propagate labels from the labeled data to the unlabeled data. By contrast, cut-paste consistency learning is more computationally efficient as it can attain strong segmentation performance after just one round of training.

\subsection{Results on CT-ICH} The five-fold cross-validation results, comprising the F1 scores and Jaccard indices (Jacc.), on the CT-ICH dataset are shown in Table \ref{tab_ich}. Different ratios of labeled examples (30\%, 50\%, 70\%) are evaluated and the 100\% setting represents the performance upper bound when all examples are labeled. Under the 30\% labeled setting, our method significantly outperforms the supervised baseline, with an improvement of 11.75\% in F1 score and 9.55\% in Jaccard index. This illustrates the benefits of cutting and pasting labeled masks onto unlabeled images when labels are scarce. Under all semi-supervised settings, cut-paste consistency emerges as the best-performing method in terms of Jaccard indices, while having competitive performance in terms of F1 score under the 50\% labeled setting. It is interesting to note that ClassMix consistency \cite{Olsson2021ClassMixSD} struggles to outperform the supervised baseline under the 30\% and 50\% labeled settings, despite our best attempts at tuning its weighting parameter. One reason might be that ClassMix was originally proposed for segmentation tasks with more than two classes, and the highly imbalance nature of the CT-ICH dataset with only two classes might produce confusing and noisy targets for the consistency learning objective. Under the 70\% labeled setting where there is sufficient labeled samples, ClassMix recovers its semi-supervised learning performance and ranked second in terms of F1 scores and Jaccard indices. Meanwhile, the proposed cut-paste consistency learning method consistently produces large performance improvements across different ratios of labeled samples.

\begin{figure*}
\begin{center}
\includegraphics[width=\textwidth]{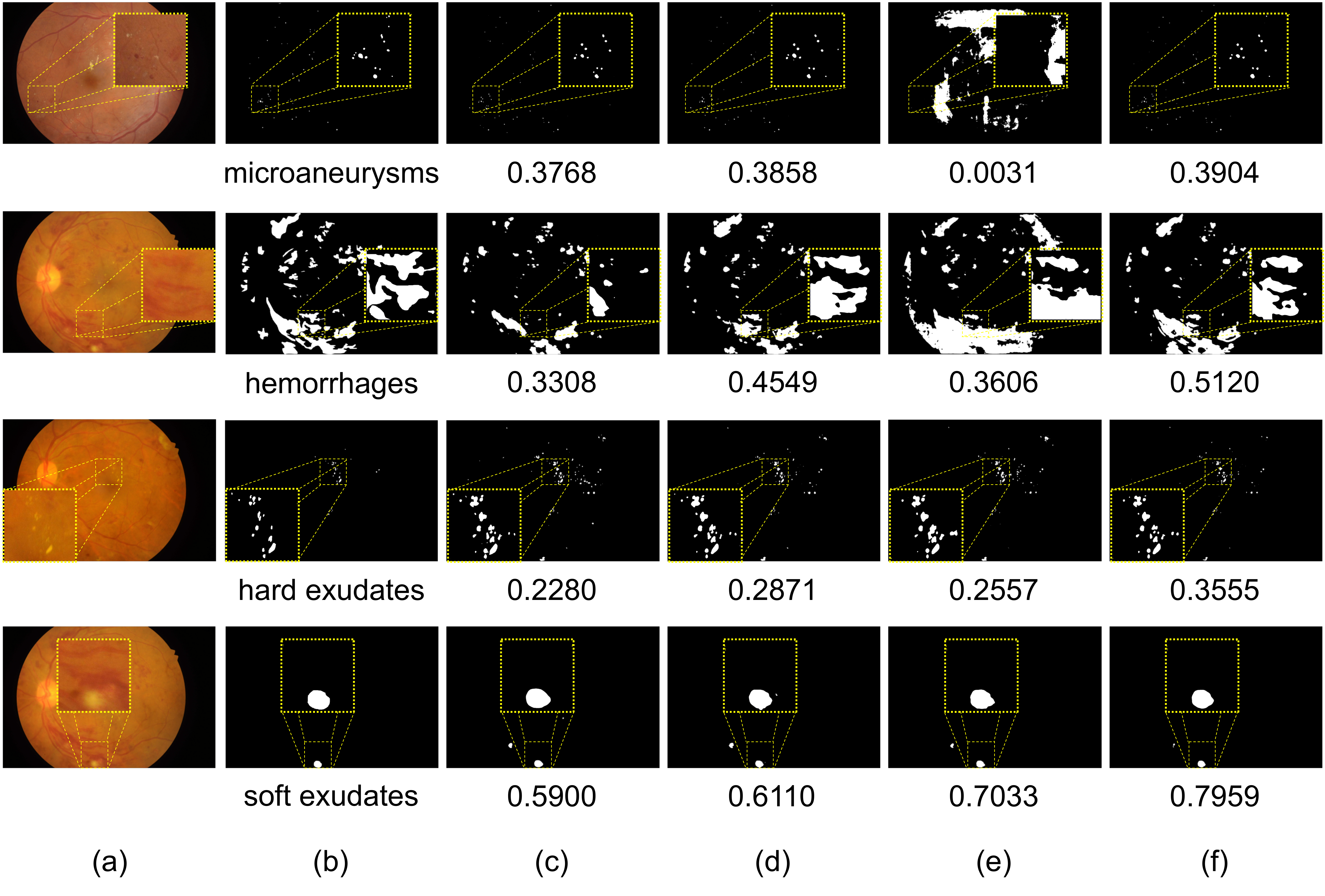}
\end{center}
\caption{Visualization of the segmentation results on IDRiD. From left to right columns: (a) input images, (b) ground truth segmentation masks, (c) predictions from the fully supervised baseline, (d) predictions from the CutMix baseline \cite{french2020semi}, (e) predictions from the ClassMix baseline \cite{Olsson2021ClassMixSD}, and (f) predictions from the proposed method. The type of lesion is indicated under each ground truth mask while the intersection over union between the ground truth mask and the predicted mask is provided under each predicted mask.}
\label{qualitative}
\end{figure*}

\subsection{Ablation Study} To investigate the impact of different components introduced in the cut-paste consistency learning method, ablation studies are conducted on the hard exudates (EX) and microaneurysms (MA) segmentation tasks in IDRiD. The comparisons are shown in Table \ref{tab_ablation}. Starting with a plain random cut-paste baseline (first row), components such as mask blurring, background blurring, color matching and consistency regularization are gradually added. As more components are added, the performance on EX segmentation steadily increases. The largest relative performance gain was observed when the background consistency term is added (3.23\% improvement comparing the fifth row to the eighth row). This is the default setup used in the main experiments. On the sixth row, another formulation of cut-paste consistency learning based on enforcing whole-image consistency is studied. This formulation simply treats the pasted objects as distractors and is equivalent to passing the synthetic images to both networks of Mean Teacher model. Although this formulation outperforms the vanilla Mean Teacher model, its performance is worse when compared to those of the background consistency formulation. This shows that the asymmetric nature of the background consistency formulation is beneficial for learning from partially labeled synthetic samples. For the background consistency term, mean square error performs slightly better than the cross-entropy function (CE). Different values of \(\lambda_u\) are also tested (last three rows) and 0.01 was found to be the optimal value. On MA segmentation, the results are generally stable across different configurations, with the default setup achieving the best overall performance.

\subsection{Qualitative Results}

Samples of the segmentation results for the baseline and proposed methods on the IDRiD dataset are compared in Fig. \ref{qualitative}. In general, the proposed cut-paste consistency learning method produces predictions with lesser number of false positives compared to other baselines. A failure case can be observed from the ClassMix consistency baseline on the segmentation of microaneurysms (row 1, column 5). The quantitative results from Table \ref{tab_idrid} shows that ClassMix struggles to reach the performance of the supervised baseline. This might be due to the inherent difficulty of this task, as evidenced by the lower AUC-PR scores across all methods in Table \ref{tab_idrid}. In contrast to other lesion types, microaneurysms are extremely small, typically only a few pixels wide, which can be challenging for ClassMix as it uses pseudo-labels to mix two images together. For extremely small objects like microaneurysms, the pseudo-labels at the first few iterations of the training stage may consist of large portions of false positives and overwhelm the true positive cases. Our proposed method does not suffer from this issue, and is shown to perform competitively in many lesion types of different shapes and sizes.

\section{Discussion}
The generation of the partially labeled synthetic samples can be viewed as an oversampling strategy for the positive instances in the input space. This is particularly suitable for lesion segmentation tasks, where lesions are typically small and sparsely distributed over the regions of interest. Pasting positive instances on top of matching unlabeled images exposes existing foreground objects to different background scenes, which effectively increases the diversity of training samples and helps boost the generalization capability of segmentation networks, as demonstrated in the experiments. Furthermore, the introduction of the background consistency regularization term helps to curb the false negative cases in the unlabeled regions of the synthetic samples. By combining them together, the proposed cut-paste consistency learning method is effective in reducing the annotation cost for lesion segmentation tasks.

It is noted that the synthesis process might not be applicable to the segmentation of large regular body structures that typically occupy a fixed location relative to other objects, such as the segmentation of lungs and livers. The generated samples will contain multiple occluding organs, which are not representative of real images and will likely cause degradation in segmentation performance. Likewise, the proposed training objective might not be suitable for panoptic segmentation tasks (e.g., the Cityscapes benchmark \cite{Cordts2016Cityscapes}) because of the diversity in sizes and types of objects that can appear in the background. Nevertheless, we believe that the proposed method could serve as a strong baseline in other tasks involving semi-supervised segmentation or detection of small irregular objects. Some examples include the segmentation of manufacturing defects from industrial photos, and detection of small targets from remote sensing images.


\section{Conclusion}
A consistency-based semi-supervised learning method for lesion segmentation is presented in this work. By utilizing lesion masks from labeled images to generate partially labeled synthetic samples and enforcing background consistency, our simple but effective cut-paste consistency method achieves significant improvements on two benchmark datasets involving fundus images and brain CT scans. This method is architecture-agnostic and could be incorporated easily into other training frameworks. For future work, one potential direction would be improving the image synthesis component using advanced techniques such as context modelling \cite{dvornik2018modeling} and image harmonization \cite{Zhu2015LearningAD, tsai2017deep}.

\section*{Acknowledgements}
\noindent The computational work for this article was fully performed on resources of the National Supercomputing Centre, Singapore (https://www.nscc.sg).

{\small
\bibliographystyle{ieee_fullname}
\bibliography{bibtex.bib}
}

\end{document}